# SMART: A Machine Learning and Monte Carlo Framework for Rapid Analysis of Stochastic Transistor Aging and Process Variation in Digital Circuits

Arash Esshaghi[1], Siavash Es'haghi[2*], Gholamreza Shahabadi[3], Alireza Moradi[4]

[1] Department of Optimization and Mathematics, Payame Noor University, Tehran, Iran.; arash.esshaghi@gmail.com
[2] Department of Electrical and Computer Engineering, SR.C., Islamic Azad University, Tehran, Iran.; s_esshaghi@iau.ac.ir
[3] Department of Electrical Engineering, National University of Skills (NUS), Tehran, Iran; ghr.shahabadi@gmail.com
[4] Department of Electrical Engineering, Se.C, Islamic Azad University, Semnan, Iran; Alireza.moradi@iau.ac.ir

[*] Corresponding Author: s_esshaghi@iau.ac.ir

**Abstract** – As CMOS technology scales into the deep nanometer regime, digital circuit reliability is increasingly threatened by the combined stochastic effects of Bias Temperature Instability (BTI) and Process Variation (PV). Traditional reliability analysis methods, which rely on computationally intensive simulations or extensive lookup tables, fail to scale efficiently for large designs, creating a critical bottleneck in design space exploration. To address this, we propose SMART, a novel framework that integrates Machine Learning (ML) with Monte Carlo simulation to enable rapid, high-fidelity reliability analysis. SMART employs Random Forest regression to predict gate delay distributions directly, bypassing time-consuming atomic model parameter extractions. Crucially, the model utilizes Bayesian Optimization for automated hyperparameter tuning, ensuring maximum predictive robustness across diverse libraries. Experimental validation on ISCAS85 benchmark circuits demonstrates that SMART achieves a 94.54% reduction in analysis time compared to state-of-the-art methods, while maintaining a remarkable average accuracy error of just 1.63%. By shifting computational complexity to an offline training phase, the proposed framework offers a scalable, accurate solution for designing resilient, reliability-aware digital systems.

## 1. Introduction

The continuing scaling of CMOS technology into the nanometer regime has enabled unprecedented levels of integration and performance in digital systems. However, this advancement comes at the cost of heightened reliability challenges, primarily due to transistor aging and process variation (PV) in nanoscale CMOS circuits [1, 2]. Transistor aging, a temporal phenomenon, manifests as gradual degradation of transistor characteristics, notably the threshold voltage, leading to increased gate delays, timing violations, and eventual circuit failure [3]. Among the aging mechanisms, Bias Temperature Instability (BTI), encompassing Negative BTI (NBTI) in PMOS transistors and Positive BTI (PBTI) in NMOS transistors, is particularly detrimental due to its pronounced impact on advanced technology nodes [4, 5]. Concurrently, process variation, a spatial effect arising from manufacturing non-idealities such as lithography imperfections, introduces stochastic deviations in transistor parameters, adversely affecting circuit yield and performance immediately after fabrication [6, 7]. The stochastic nature of both PV and BTI in sub-45nm technologies exacerbates these challenges, necessitating robust analysis methods to ensure circuit reliability and lifetime.

The simultaneous analysis of stochastic aging and PV (APV) is critical for modern digital circuit design. These phenomena directly impact key metrics such as yield, performance, and operational lifetime, which are important for applications ranging from consumer electronics to automotive systems. For instance, PV can reduce fabrication yield by causing deviations in transistor characteristics, while aging-induced delay increases can lead to timing failures over time, compromising system lifetime [8].

As digital design shifts toward higher levels of abstraction to manage complexity, there is an urgent need for fast, accurate models that enable reliability-aware design space exploration [9]. Such models must account for the stochastic behavior of BTI and PV while maintaining computational efficiency to support iterative design optimization.

Existing methods for analyzing transistor aging and PV, while valuable, fall short in addressing the demands of large-scale circuits and high-level design. For example, the statistical framework proposed in [10] characterizes reliability for small circuits but lacks scalability to complex designs due to its computational intensity. Similarly, the method in [11] offers a comprehensive analysis of stochastic aging and PV using lookup tables (LUTs) and Monte Carlo simulations but suffers from long execution times, making it impractical for rapid design space exploration. Commercial tools like MOSRA and RelExpert provide circuit-level reliability analysis but are often time-consuming and tailored to lower abstraction levels. These limitations highlight a critical research gap: the absence of a fast, accurate, and scalable method for simultaneous analysis of stochastic APV in modern digital systems, particularly one that efficiently optimizes machine learning (ML) models for predictive accuracy in high-dimensional parameter spaces.

Bayesian Optimization (BO) emerges as a transformative approach for hyperparameter tuning in machine ML-based aging prediction, offering superior efficiency over traditional methods such as grid search or random search coupled with k-fold cross-validation. While k-fold validation provides robust model assessment by partitioning data into folds for repeated training and testing, it becomes computationally prohibitive in exhaustive searches across vast hyperparameter grids, often requiring thousands of evaluations without intelligent guidance [12]. In contrast, BO employs probabilistic surrogates, typically Gaussian processes, to model the objective function and an acquisition function to strategically select promising hyperparameters, minimizing evaluations while maximizing performance in black-box optimization scenarios [13]. This efficiency is particularly impactful in transistor aging modeling, where BO has been leveraged for analog circuit sizing under degradation effects [14], power transistor design with constraints on reliability metrics [15], and failure prediction in insulated-gate bipolar transistors (IGBTs) via optimized feature extraction [16]. By integrating BO with evolutionary algorithms or neural networks, recent works achieve accelerated convergence in reliability-aware optimization, surpassing traditional k-fold-based tuning in sample efficiency and scalability for nanoscale devices [17, 18].

We propose a novel methodology that integrates ML and Monte Carlo techniques to model and analyze the effects of stochastic aging and PV in standard cells. Our approach leverages Random Forest (RF) regression, enhanced by BO for automated hyperparameter tuning, to predict gate delay distributions, eliminating the need for time-intensive atomic model parameter extraction during analysis. By combining this ML-based model with a Monte Carlo framework, we enable rapid and accurate reliability analysis at higher abstraction levels, facilitating efficient design space exploration. The contributions of this work are as follows:

- Development of an ML-based model for predicting standard cell delay distributions, accounting for stochastic APV, with automated hyperparameter tuning via BO to enhance robustness and accuracy over traditional grid search methods.
- A Monte Carlo-based framework for simultaneous analysis of APV, ensuring scalability to large circuits while incorporating probabilistic optimization for efficient exploration of design spaces.
- Significant performance improvement, achieving a 94.54% reduction in analysis time with a 1.63% accuracy error compared to state-of-the-art methods, validated on ISCAS85 circuits and demonstrating superior efficiency in reliability-aware design.

The remainder of this paper is organized as follows: Section 2 reviews related work in reliability modeling; Section 3 details the proposed SMART methodology and BO integration; Section 4 presents extensive experimental results validating the framework's accuracy and speed; and Section 5 concludes with a discussion on future directions.

## 2. Related works

The aging of transistors is significantly influenced by the BTI phenomenon, particularly in technologies below 45 nm [4]. Over the years, various theories have been developed to predict and explain its effects, with the Reaction-Diffusion (RD) and Hole-Trapping models being among the most prominent [2]. Based on these theories, several models and equations have been proposed for analyzing aging impacts [19]. However, the complexity of these models often limits their applicability to the device level or small circuits at the gate/cell level.

Recent advancements have introduced hierarchical frameworks for analyzing NBTI effects on digital circuit performance, considering factors such as voltage variations and input switching activity [20]. Commercial tools like MOSRA and RelExpert enable circuit-level reliability analysis, but their application to large circuits remains time-consuming. Static Timing Analysis (STA) offers a faster alternative for large circuits by using gate models instead of full simulations [21], with aging-aware STA incorporating degradation effects into these models [22]. LUTs have been employed to model aging behavior in gates [21, 23], though these approaches requiring multiple LUTs increase complexity [23]. To mitigate this, sensitivity-based methods extract gate delay responses to threshold voltage variations, reducing LUT requirements [24]. This method reduces the number of LUTs and simplifies the modeling process, thereby reducing the computational burden. However, the accuracy is compromised.
With CMOS scaling into the deep nanometer regime, BTI exhibits stochastic behavior due to reduced defect counts and increased randomness [25]. Atomic models describe this stochastic BTI at the transistor level [26], but their complexity restricts use to small gates. Advances have extended atomic models to standard library cells via statistical STA [27], though these are time-intensive and often limited to critical paths, neglecting PV. A concurrent analysis framework for stochastic aging and PV was introduced in [11], utilizing LUTs to characterize cells and extracting atomic parameters via logical simulation and Monte Carlo. However, this approach suffers from high execution times caused by the heavy computational overhead of atomic model parameters extraction and extensive LUT interpolations. [28] proposes an ML-based model to predict stochastic BTI effects, using Monte Carlo simulations to generate training data and evaluating ensemble algorithms like RF, achieving considerable runtime reduction with up to 98% accuracy. However, this method neglects PV.
BO has emerged as a transformative technique to enhance the efficiency of ML-based aging analyses, particularly in navigating the high-dimensional parameter spaces inherent to reliability modeling. Traditional ML approaches, such as those relying on manual tuning with k-fold cross-validation, often require exhaustive grid searches across hyperparameter sets, leading to significant computational overhead and suboptimal configurations [12]. In contrast, BO leverages probabilistic surrogates, typically Gaussian processes, and acquisition functions to intelligently sample the search space, reducing the number of evaluations while optimizing model performance. For instance, [13] proposed a batch-constrained BO approach for analog circuit synthesis, leveraging multi-objective acquisition ensembles to efficiently explore design spaces while minimizing costly evaluations. This method achieves significant speedups in optimizing performance metrics under variability, though primarily for analog contexts. [17] applied BO to area optimization of op-amps, demonstrating faster convergence over traditional methods, but it struggles with complex multi-objective problems involving aging. Hybrid approaches integrate BO for hyperparameter tuning within evolutionary algorithms, dynamically optimizing parameters like population size to enhance search efficiency in aging mitigation for FinFET cells [18].
Recent ML-based aging prediction and modeling methods further advance scalability. [12] introduced ML-TIME, an ML-driven timing analysis framework that models propagation delays under PVs and aging-induced $V_{th}$ shifts using feedforward MLPs tuned via BO-based neural architecture search. It bridges HSPICE and STA simulations for accurate path-level predictions in FinFET PDKs, offering an extraordinary speedup over SPICE. Similarly, [18] proposed RelOps, utilizing LightGBM models for delay and power prediction in 16nm FinFET cells, accounting for PVT over 10 years, and integrating a hybrid evolutionary algorithm with Bayesian-tuned hyperparameters to optimize transistor sizing, yielding up to 36.97% PDP improvement without aging and 34.94% with aging. [29] introduced efficient learning strategies to reduce data requirements for ML-based characterization of aging-aware standard cell libraries, achieving up to 0.93 $R^2$ accuracy with fewer SPICE simulations, though without full quantification of absolute error metrics essential for optimization workflows.
In the realm of High-Level Synthesis (HLS), efforts to integrate reliability considerations have gained traction [30]. [31] introduced a novel framework aimed at embedding stochastic APV considerations into functional unit (FU) binding and port assignment during HLS. This approach utilizes an RF model with 98.7–98.9% accuracy to predict workload-induced APV effects, employing Simulated Annealing (SA) optimization to concurrently minimize degradation and interconnection costs while ensuring uniform FU degradation.
With growing circuit complexity, higher abstraction levels necessitate fast, efficient methods for joint stochastic APV analysis. Our work addresses this by combining ML models with Monte Carlo, enabling rapid reliability assessment while surpassing prior approaches in speed and accuracy.

### 2.1 Atomic Model of Transistor Stochastic Aging

The stochastic nature of the BTI phenomenon at the transistor level is explained by the atomic model introduced in [25]. This model enables calculation of the total change in the threshold voltage of each transistor arising from BTI. The BTI phenomenon involves two phases of stress and recovery, resulting from defects in the oxide gate that are charged and discharged, respectively [32]. The charging and discharging of each defect during the capture and emission time have a distinct effect on the threshold voltage of the transistor. Furthermore, the impact of each occupied defect on $v_{th}$ follows an exponential distribution, Equations (1) and (2).

$$\Delta v_{th} \sim Exp(\eta) \quad (1)$$

$$\eta \propto \frac{1}{(LW)} \quad (2)$$

Where $\eta$ represents the average effect of each defect, experimentally extracted [33, 34].

The average number of defects present in each transistor, $N_T^{avg}$, is determined in Equation (3), utilizing the density of defects, $n_T$. This density is obtained from capture/emission time (CET) maps ( $f_{CET}$ ) presented with experimental data collected in [34], as illustrated in Equation (4). The distribution map of the capture time ( $\tau_C$ ) and emission time ( $\tau_e$ ) depicts the CET map.

$$N_T^{avv} = W.L.n_T \quad (3)$$

$$n_T = \iint f_{CET}(\tau_C, \tau_e) d\tau_c d\tau_e \quad (4)$$

Only a portion of existing defects contribute to BTI-induced degradation of the threshold voltage of the transistor, Equation (5). To establish the average number of active defects, it is essential to calculate the probability of each defect being occupied ( $p_{OCC}$ ), as shown in Equation (6). Then the active effects ratio is computed using Equation (7).

$$N_T = \rho.N_T^{avv} \quad (5)$$

$$p_{OCC} = \frac{1-e^{-\frac{sp}{f\tau_c}}}{1-e^{-\frac{1}{f}(\frac{sp}{\tau_c}+\frac{1-sp}{\tau_e})}}(1-e^{-t_{steress}(\frac{sp}{\tau_c}+\frac{1-sp}{\tau_e})}) \quad (6)$$

$$\rho = \frac{\iint f_{CET}(\tau_C, \tau_e) P_{OCC}(\tau_c, \tau_e, sp, t_{steress}, f) d\tau_c d\tau_e}{\iint f_{CET}(\tau_C, \tau_e) d\tau_c d\tau_e} \quad (7)$$

In the Equation (7), $f$ denotes frequency, and $t_{steress}$ represents the total stress time. It is notable that the probability of filling defects depends on the characteristics of the signal probability (SP) and CET of defects. By employing the average number of defects in each transistor, the distribution of the number of defects is determined to follow a Poisson distribution with the average of $N_T$, as expressed in Equation (8).

$$n \sim poiss(N_T) \quad (8)$$

Using the exponential distribution, the total $v_{th}$ change can be calculated by summing the effects of all defects.

The atomic model of NBTI stochastic behavior analysis is complex and only applicable at the transistor level; therefore, it cannot be used for higher abstraction levels in design. However, various methods have been proposed to detect aging in large circuits based on this model. In [35], the aging of diverse architectures of 32-bit adder circuits was evaluated by combining these models and commercial tools. Utilizing STA and the Monte Carlo method, [25] proposed a method for simultaneous analysis of the effects of stochastic APV. In addition, aging effects were also determined through STA in [36, 37]. Although approaches produced by the aforementioned methods operate at relatively low speeds, the severity of stochastic aging was evaluated based on ML in [28]. However, this approach evaluates the aging of circuits using existing methods, and generating the training set is a time-consuming task.

## 3. Proposed SMART Methodology

This section outlines the various steps involved in our proposed approach.

### 3.1 Characterizing Standard Cells

The delay of each gate is dependent on its internal structure, input transition time, and load capacitor ( $C_L$ ). Usually, the delay values corresponding to various values of load capacitor and input transition time are stored in 2D LUTs that can be used to determine the delay of new circumstances via interpolation. The commercial tool Liberate from Cadence can be used to generate the LUTs for standard cells. Alternatively, we can extract the cell characteristics through extensive simulations in HSPICE. To accomplish this, we calculate the gate delay for several different combinations of load capacitor and input transition time values, and store the results in the LUTs. In order to obtain the delay of a circuit using STA, we must also obtain the output transition time based on the working conditions, which includes the load capacitor and input transition. To achieve this, we simulate the output transition time of each gate/standard cell for various values of load capacitor and input transition time, and store the results in the LUTs.
Equation (9) is then utilized to interpolate the zero time delay (fresh delay) and output transition time for an arbitrary value of load capacitor and input rise time ( $T_R$ ).

$$\begin{aligned} D_0 &= f_1(C_L, T_R) \\ t_{r_out} &= f_2(C_L, T_R) \end{aligned} \tag{9}$$

Here, delay values and output transition values are obtained through interpolation performed by $f_1$ and $f_2$ . To achieve this, the desired load capacitor and input transition time values are used with LUTs.
In order to account for APV effects, we obtain gate delay values by applying changes in threshold voltage to transistors and storing the resultant delays in LUTs. To obtain the degraded delay in each state through simulation, various combinations of threshold voltage changes can be considered for all gate transistors at the same time, such as in [11].
To enhance precision and account for these simultaneous changes, a set of LUTs is created for each gate, with each corresponding to a possible combination of simultaneous shifts in threshold voltage values ( $\Delta vth$ ) of all transistors within that gate. For instance, for the INV_X1 gate, numerous LUTs labeled as $INV_X1_{\Delta vthn \Delta vthp}$ are generated, each with a specific value for $\Delta vthn$ (related to NMOS) and $\Delta vthp$ (related to PMOS). Consequently, timing information of a gate (including rising/falling delays and transition times) corresponding to different combinations of threshold voltage variations of its transistors is stored in separate LUTs. To manage the number of LUTs, discrete values are used for $\Delta vth$ of transistors, and for a specific combination of $\Delta vth$ values.

### 3.2 Analysis of stochastic aging and PV of standard cells

In deterministic circumstances, the threshold voltage degradation of a transistor caused by BTI under a known SP vector has a definite value. In such a scenario, the gate delay degradation resulting from aging phenomena is determined deterministically as a function of the SP vector (the SP values of the gate inputs), the vector of $T_R$ (the transition times of the inputs), and the $C_L$ (load capacitor). However, BTI exhibits a stochastic nature in deep nano-scale technologies. This indicates that identical transistors operating under same conditions, including the SP vector, exhibit varying threshold voltage degradations, yielding stochastic circuit delays.
When given an SP vector, the number of defects in the transistor gate oxide is obtained statistically according to a Poisson distribution outlined in Equation (8). Each defect's charging and discharging additionally possesses a stochastic effect on the threshold voltage degradation of the transistor, which follows an exponential distribution demonstrated in Equation (1). The total threshold voltage degradation of the transistor *m*, attributed to the NBTI phenomenon, is equal to the sum of the impacts of all active defects, as expressed in Equation (10):

$$\Delta v_{th,m}^{NBTI} = \sum_{k=1}^{NT} \Delta v_{th,k} \tag{10}$$

In Equation (10), $NT$ represents the number of active defects obtained statistically, while $\Delta v_{th,k}$ denotes the contribution of defect *k* to the transistor *m*'s threshold voltage degradation caused by charging of the defect *k*. This value is obtained by sampling from the exponential distribution.

Finally, Equation (11) indicates that the total change of the transistor $m$'s threshold voltage is the sum of the effects of NBTI ($\Delta v_{th,m}^{NBTI}$) and PV ($\Delta v_{th,m}^{pv}$). It is possible to derive the threshold voltage change that results from PV through a normal distribution with a mean of zero and a standard deviation obtained by Pelgrom's law.

$$\Delta v_{th,m} = \Delta v_{th,m}^{NBTI} + \Delta v_{th,m}^{pv} \qquad (11)$$

### 3.3 Aging-aware ML -based gate delay model

Generating a sufficiently large training set that includes the features of (SP, $C_L$, $T_R$) and targets including the average values (μ) and standard deviation (σ) of the gate delay can be obtained by applying various values of the features. This approach enables the creation of an aging-aware ML-based gate delay model for each standard cell. By utilizing this training set, the parameters of gate delay distribution (μ and σ) can be predicted accurately and quickly using a suitable trained ML algorithm. For instance, RF is an optimal model for this task due to its speed, accuracy, and flexibility. Recent research [28] demonstrates that the RF model is capable of modeling aging effects effectively. The process is depicted in Figure 1.

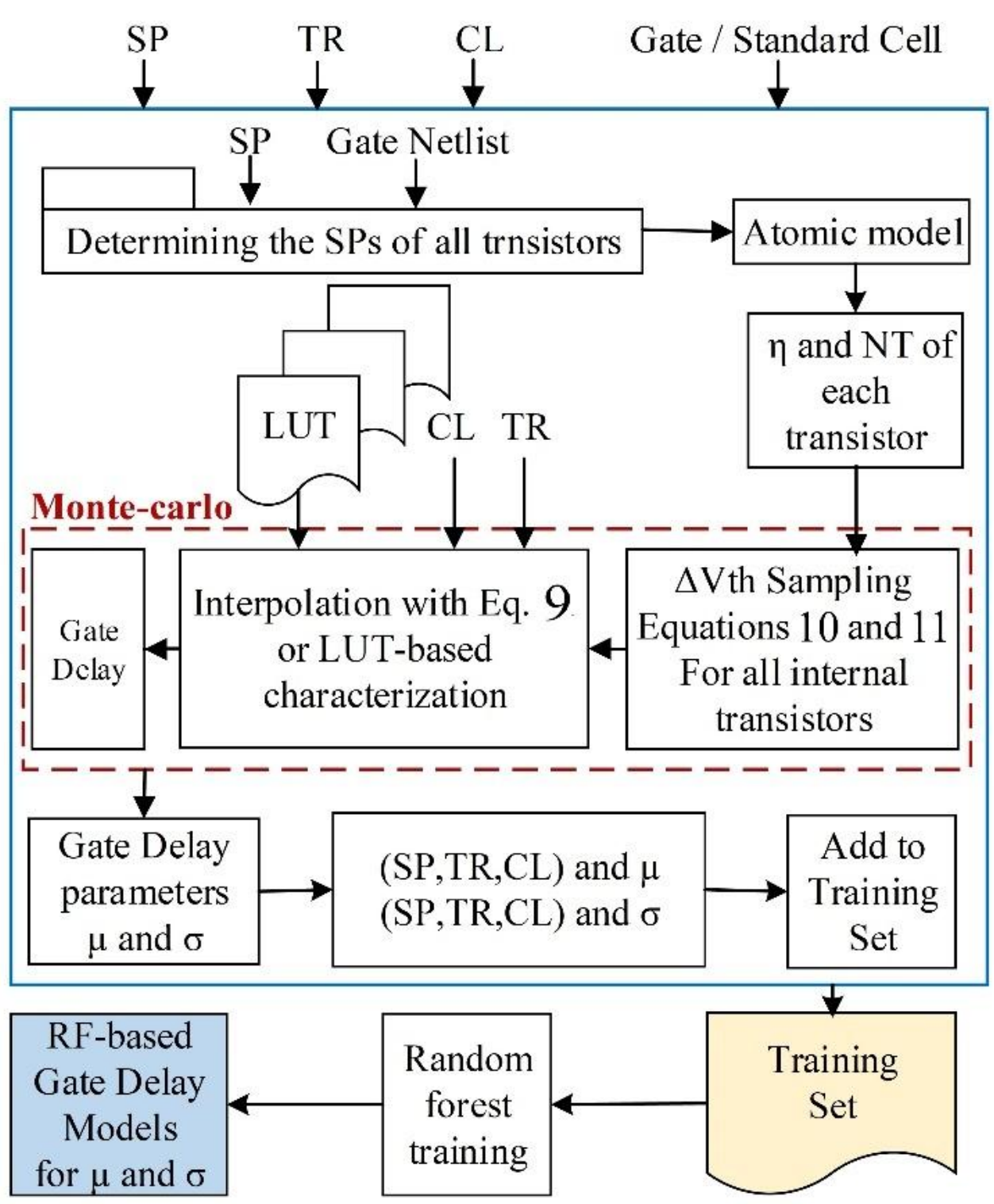

**Fig. 1.** The flow of developing the aging-aware machine-learning based model for gate delay

Developing a ML model to predict gate delay (μ and σ) is central to our method, significantly boosting the efficiency of digital circuit timing analysis. By bypassing the need to extract atomic model parameters during analysis, a notoriously time-intensive step (as seen in methods like [11]), our approach dramatically reduces execution time. Furthermore, it eliminates the computationally expensive interpolation of LUT values for gate delay, especially for complex gates. This RF-based model streamlines the entire process, with detailed explanations provided in the next subsection.

On the other hand, it should be noted that developing these ML models requires creating a large training set for each standard cell. To achieve this, two training sets in the form of $\{((SP_i, T_{R,i}, C_{L,i}), \mu_i), ...\}$ and $\{((SP_i, T_{R,i}, C_{L,i}), \sigma_i), ...\}$, respectively for training the ML model for regression of μ and σ of each gate, need to be generated, as illustrated in Figure 1. For this purpose, firstly, the atomic model parameters of each transistor corresponding to the desired SP are determined. Then, Monte Carlo simulation with $\Delta v_{th}$ sampling (using equations 12 and 13) is performed, and in each Monte Carlo iteration, the gate delay is calculated using the timing characteristics of the gate (leveraging LUTs) and the given $T_R$ and $C_L$. At the end of the Monte Carlo simulation, by aggregating the obtained delays in various iterations, the μ and σ values of the gate delay distribution are determined statistically. This process is repeated for various values of SP, $T_R$, and $C_L$ for each gate, and ultimately, the training set is created.
With these explanations, it is evident that the application of the atomic model in the proposed method is to generate a training set for training the RF model. This process is an offline process and does not play a role in the execution time of digital circuit timing analysis with the proposed method. In fact, after developing this model, the atomic model is no longer used in circuit timing analysis. It should be noted that standard cells are much smaller compared to practical circuits, and their analysis (for each specific SP) using atomic model does not pose much challenge even with a large number of iterations.
**Automated Hyperparameter Optimization:** To ensure the RF regression model achieves optimal predictive accuracy and robustness for gate delay distributions under stochastic APV, we employ BO for automated hyperparameter tuning [38]. Unlike traditional grid or random search methods, which exhaustively evaluate numerous hyperparameter combinations and incur significant computational overhead, BO leverages a probabilistic surrogate model to intelligently explore the hyperparameter space. By modeling the relationship between hyperparameters and model performance, BO iteratively selects configurations that maximize predictive accuracy, minimizing the number of evaluations required. This approach is particularly effective in the context of transistor aging, where complex, high-dimensional parameter spaces demand efficient optimization strategies. Conducted offline, BO ensures no runtime penalty during circuit analysis, preserving the framework's speed advantage while enhancing the RF model's ability to generalize across diverse standard cell libraries and operating conditions. This results in a robust, high-fidelity model that significantly improves the accuracy and scalability of reliability-aware timing analysis.

### 3.4 Determining the Delay Distribution of the Digital Circuit

Calculating the delay distribution of a digital circuit is crucial to evaluate its overall statistical performance. One commonly used method for this calculation is the Monte Carlo method, in which a representative sample from the distribution of each gate's delay is used for every iteration. Utilizing the proposed ML-based gate delay model, the statistical parameters of each gate's distribution are obtained based on its workload. The proposed workflow is demonstrated in Figure 2.

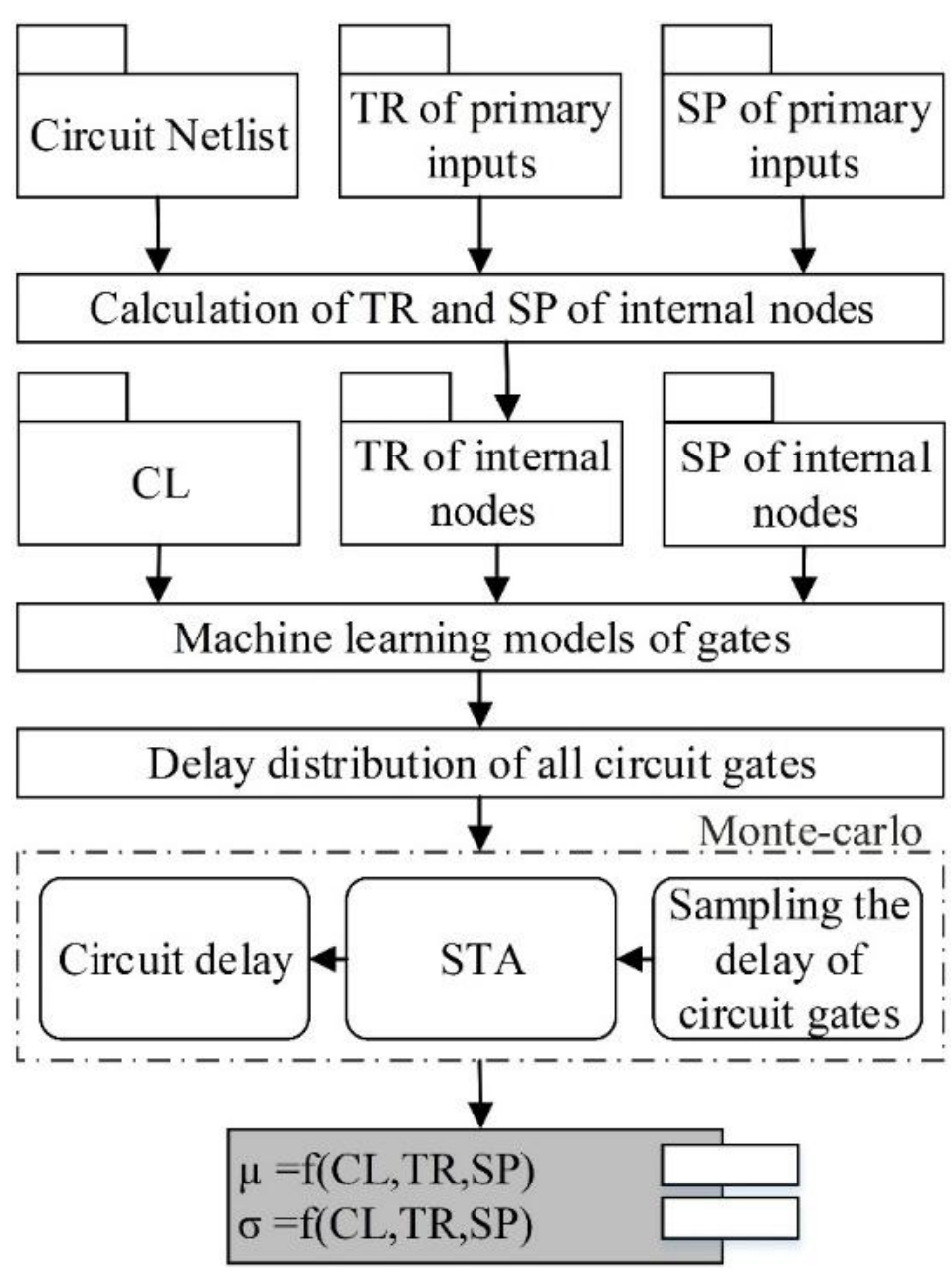

**Fig. 2.** The flow of the proposed approach for determining the delay distribution of the digital circuit

The details of the proposed method are as follows: by having the SP values of primary inputs (primary SP vector), the SP values of all internal gates of the circuit are calculated either through behavioral simulation or using logic simulators. Additionally, the load capacitance ($C_L$) of each circuit's gate is determined based on the capacitances of the gates connected to its output. Furthermore, with the $T_R$ of the primary inputs (primary $T_R$ vector), the input $T_R$ values of all internal gates of the circuit are calculated using the relevant timing transition tables (gathered in the standard cell characterization stage). This way, the SP, $T_R$, and $C_L$ values for all circuit gates are obtained. With these values, developed ML models can predict the distribution characteristics (μ and σ) of delay for all internal gates.
Having μ and σ for the delay of all gates, a set of delay samples for each gate can be generated, representing suitable representatives for the gate delay according to its distribution.
Then, Monte Carlo simulation is performed in such a way that in each iteration, samples of all gates' delays are selected based on their respective delay distributions. By using these values and performing STA, the circuit delay corresponding to the selected samples in that iteration can be obtained. With sufficient iterations, a set of delay values for the circuit is obtained, and by aggregating them, the μ and σ values of the circuit's delay can be calculated.
The key advantage of the proposed method lies in directly sampling gate delays according to the delay distribution characterized by parameters predicted by the ML-based model. In contrast, the method in [11] recalculates transistor threshold-voltage shifts in every iteration using Equations 10 and 11 based on atomic model parameters (computed once at the beginning), and then evaluates circuit delay through look-up tables (LUTs) and static timing analysis (STA). This fundamental difference in the analysis approach enables the proposed method to achieve significantly better runtime performance than the method presented in [11].

## 4. Experiments

### 4.1 Setting up the experiments

To validate the efficacy of the proposed SMART framework for analyzing stochastic A PV, a comprehensive experimental setup was established, focusing on standard cell libraries and benchmark circuits. A standard cell library was produced using 7-nm technology by scaling an existing open-source 45-nm library. These libraries included a diverse set of gates, ranging from simple NAND/NOR gates to more complex structures like AOI and OAI gates, ensuring a robust representation of typical digital circuit components.
The standard cells were characterized using HSPICE with BSIM4 models to simulate transistor behavior under various operating conditions. The characteristics of standard cells were first extracted and stored in LUTs. For this purpose, netlists were produced for each standard cell, with four different values of load capacitor ($C_L$) and four different values for input transition time ($T_R$), which were then simulated with HSPICE. The netlists were automatically generated using a Python script, and the resulting data was stored. This approach enabled the production of LUTs that were used to extract the zero time delay (fresh delay) and output transition time of standard cells associated with desired $C_L$ and $T_R$. Additionally, a set of netlists was produced for each cell by applying different combinations of $\Delta v_{th}$ to its transistors for each $C_L$ and $T_R$ combination, which were then simulated with HSPICE to produce the LUTs of gate delay timing information corresponding to changes in the threshold voltage of transistors. The results were read and stored using a Python script, resulting in timing information LUTs. In this way, we implement the LUT-based method for characterizing standard cells.
Atomic model parameters, including η and $N_T^{avv}$, were calculated for each transistor of a standard cell based on its W and L. After determining the value of the input SP vector, the average number of active defects ( $N_T^{avv}$ ) was determined, which was the mean of the Poisson distribution of active defects of the transistor. A set of 3000 samples for the number of active defects was then produced for each transistor by sampling the associated distribution. As a result, a set of 3000 samples of threshold voltage variations due to APV was produced for each transistor using Equation (11). During each iteration of the Monte Carlo simulation of a standard cell, a sample of $v_{th}$ variation was applied to every transistor threshold voltage, resulting to a combination of threshold voltage shifts for its internal transistors, and the corresponding gate delay is determined by leveraging the LUTs. Finally, the Monte Carlo simulation for each standard cell was completed with 3000 iterations, and the statistical parameters of the delay distribution were determined, namely the mean (μ) and standard deviation (σ). In fact, with given SP, $T_R$ and $C_L$, and after the completion the Monte Carlo simulation, the corresponding μ and σ are obtained.
Lastly, various simulations were performed (for each standard cell) using 2000 different values of $C_L$, $T_R$, and SP, resulting in the calculation of μ and σ for each combination. The values of $C_L$,$T_R$, and SP were used as the features, with the corresponding μ and σ serving as the targets. This constituted the training set, which was then used to train a RF regression model. The ML model development leveraged Scikit-learn for RF implementation, NumPy for data processing.
To enhance the predictive accuracy of the RF regression model for gate delay distributions under APV, automated hyperparameter optimization was implemented using BO with Optuna, a robust Python library for efficient hyperparameter search. Optuna was configured to optimize key RF hyperparameters, including n_estimators (100–1000, step size 100), max_depth (10–50, step size 5), min_samples_split (2–10, step size 1), and min_samples_leaf (1–5, step size 1), within a comprehensive search space tailored to the standard cell library. BO utilized a Gaussian process as the probabilistic surrogate model to predict model performance, coupled with an acquisition function to intelligently select hyperparameter configurations. The optimization process, conducted offline over 100 trials, employed Optuna's Tree-structured Parzen Estimator (TPE) sampler and median pruning to minimize computational overhead while maximizing accuracy. Five-fold cross-validation was integrated to ensure model generalization, reducing overfitting across diverse operating conditions.

In order to determine the delay of each benchmark circuit, a gate level description was obtained using Design Compiler. The SP of all internal nodes (inputs of the gates) associated with a specific primary SP vector of the circuit primary inputs was then calculated using the logical function of the gates, while the transition time ($T_R$) of internal nodes was determined using the transition time LUTs and the transition times of primary inputs. It is worth mentioning that logical simulators such as ModelSim can also be used to obtain the SPs of all internal nodes. The output capacitance ($C_L$) of each gate was calculated based on the connected gates.

With the SP, $C_L$ and $T_R$ values of each gate in the circuit, statistical parameters of the delay distribution of the gate, namely $\mu$ and $\sigma$, were determined using ML models. A set of 10,000 samples of delay was then generated for each gate using these parameters. For determining the delay of a circuit, Monte Carlo simulation was performed with 10,000 iterations, and in each iteration a sample delay was used for each gate delay. Block-based STA was performed to determine the circuit delay in each iteration. Although Primetime was unavailable, a Python script was written to perform block-based STA. A set of delay values for the circuit was obtained with 10,000 samples at the end of the Monte Carlo simulation, which can be used to calculate $\mu$ and $\sigma$ values of the circuit delay. For runtime comparison, the proposed method and the one proposed in [11] was implemented in Python. The proposed method was evaluated using ISCAS85 circuits, whose specifications are listed in Table 1.

**Table 1.** Specifications of ISCAS85 circuits

| Benchmark | Circuit Function | Number of gates | Number of inputs | Number of outputs |
|---|---|---|---|---|
| C432 | Priority Decoder | 160 | 36 | 7 |
| C880 | ALU and Control | 383 | 60 | 26 |
| C1355 | ECAT | 546 | 41 | 32 |
| C1908 | ECAT | 880 | 33 | 25 |
| C2670 | ALU and Control | 1193 | 233 | 140 |
| C6288 | 16-bit Multiplier | 2406 | 32 | 32 |
| C7552 | ALU and Control | 3512 | 207 | 108 |

The experiments were run as a single thread on a personal computer with an Intel Core i7 Q740 (1.73GHz, 6MB cache, first generation) processor and 4GB of RAM.

### 4.2 Experimental results

**Runtime Comparison:** The proposed method and the method of [11] were implemented, and their respective runtimes are shown in Table 2. The proposed method achieved a noticeable 94.54% reduction in analysis time compared to the state-of-the-art approach [11], which relies on atomic model parameter extraction and extensive LUT interpolations. This dramatic speedup was consistent across all benchmark circuits, with runtime measurements averaging 85 minutes for the SMART framework versus 2386 minutes for [11] on a c7552 circuit with 10,000 Monte Carlo iterations. This efficiency stems from the offline-trained RF model, which eliminates runtime parameter extraction and LUT interpolation, enabling rapid design space exploration critical for modern VLSI workflows.

**Table 2.** Runtime evaluation

| Benchmark | Runtime (sec) | | %improvement |
|---|---|---|---|
| | [11] | Proposed | |
| C432 | 17236 | 1264 | 92.67 |
| C880 | 31014 | 2078 | 93.3 |
| C1355 | 39788 | 2285 | 94.26 |
| C1908 | 36047 | 2307 | 93.6 |
| C2670 | 59029 | 2731 | 95.37 |
| C6288 | 82897 | 3196 | 96.14 |
| C7552 | 143207 | 5124 | 96.42 |
| **Average** | | | **94.54** |

**Accuracy Comparison:** Despite faster runtime of the proposed method compared with [11], the proposed method maintains high accuracy, as demonstrated by the mean values and standard deviation presented in Table 3, where we compare the proposed method and [11]. Accuracy, measured as the average relative difference between the results of the proposed method and the results of [11], was maintained at an impressive average 1.63% and 2.53%, respectively for μ and δ, across all ISCAS85 circuits. This precision was validated through extensive Monte Carlo simulations, ensuring robust reliability assessments under varying signal probabilities, transition times, and load capacitances.

**Table 3.** Evaluation of the accuracy of the proposed method

| Bench. | [11] | | Proposed | | %error | |
|---|---|---|---|---|---|---|
| | μ | δ | μ | δ | μ | δ |
| C432 | 6.43 | 0.69 | 6.32 | 0.68 | 1.71 | 1.45 |
| C880 | 5.86 | 0.81 | 5.77 | 0.83 | 1.54 | 2.47 |
| C1355 | 6.23 | 0.73 | 6.32 | 0.72 | 1.44 | 1.37 |
| C1908 | 5.89 | 0.83 | 5.79 | 0.86 | 1.7 | 3.61 |
| C2670 | 6.01 | 0.79 | 6.11 | 0.76 | 1.66 | 3.8 |
| C6288 | 6.24 | 0.83 | 6.14 | 0.81 | 1.6 | 2.41 |
| C7552 | 6.76 | 0.76 | 6.88 | 0.74 | 1.78 | 2.63 |
| **Average** | | | | | 1.63 | 2.53 |

**Comparison of Baseline RF and Tuned Model Performance:** To elevate the framework's predictive power, a dedicated comparison between the baseline RF model (tuned via manual grid search) and the optimized model (enhanced with BO using Optuna) was conducted, as detailed in Table 4. The baseline RF, configured with a 5-fold grid search, achieved the reported 1.93% and 2.76% accuracy error across the dataset, respectively for μ and δ. The tuned model, demonstrated a potential error reduction to 1.63% and 2.53%, reflecting improved generalization across diverse gate types and operating conditions. This enhancement is attributed to BO's intelligent sampling and pruning, which optimized hyperparameter configurations for stochastic aging contexts. The overall comparison of two approaches is mentioned in Table 5.

**Table 4.** Comparison of Baseline RF and Tuned Model Performance

| Bench. | % error | | | |
|---|---|---|---|---|
| | Baseline RF (Grid Search) | | Tuned RF (BO) | |
| | δ | μ | δ | μ |
| C432 | 1.45 | 1.71 | 2.9 | 2.18 |
| C880 | 2.47 | 1.54 | 2.47 | 1.71 |
| C1355 | 1.37 | 1.44 | 2.74 | 1.93 |
| C1908 | 3.61 | 1.7 | 2.41 | 1.87 |
| C2670 | 3.8 | 1.66 | 3.8 | 2.16 |
| C6288 | 2.41 | 1.6 | 2.41 | 1.92 |
| C7552 | 2.63 | 1.78 | 2.63 | 1.78 |
| **Average** | 2.53 | 1.63 | 2.76 | 1.93 |

**Table 5.** Comparison of Baseline RF and Tuned Model Performance

| **Metric** | **Baseline RF** | **Tuned RF** |
|---|---|---|
| μ Accuracy Error (%) | 1.93 | 1.63 |
| δ accuracy Error (%) | 2.76 | 2.53 |
| Runtime Impact (%) | 0 (Offline) | 0 (Offline) |
| Training Overhead | Moderate | High (100 trials) |

**Complex and large Gates and Synthesis Approaches:** It is noteworthy that the runtime gains of the proposed method are particularly prominent in circuits composed of more complex gates. To demonstrate this, we conducted two synthesis approaches with different synthesis objectives. The first approach aimed to reduce chip area (area-driven synthesis), whereas the second approach aimed to increase circuit speed (performance-driven synthesis). The circuit synthesized for performance-driven synthesis had more large and complex gates, with more transistors. Table 5 presents the runtime results for both synthesis scenarios. Remarkably, the proposed method exhibited a 2.1% faster runtime improvement in the reliability analysis of performance-driven synthesized circuit compared to the same circuit synthesized with the area-driven approach. Although the percentage improvement was relatively small (about 2.1%), the absolute values were noteworthy. The proposed method demonstrated a runtime improvement of 49,208 (52383-3175) seconds (93.9%) in area-driven synthesis and 93,936 (97837-3901) seconds (94%) in performance-driven synthesis.

**Table 6.** Evaluation of runtime for two synthesis approaches

| Bench. | Runtime (s) | | | | | |
|---|---|---|---|---|---|---|
| | area-driven synthesis | | | performance-driven synthesis | | |
| | [11] | ours | Impv. | [11] | ours | Impv. |
| C6288 | 52383 | 3175 | %93.9 | 97837 | 3901 | %96.0 |

**Offline Training Overhead:** It is noteworthy that achieving this significant improvement required a considerable investment of time spent on generating the training sets and training the ML-based model (RF regression). However, this training was done offline and only once, without affecting the runtime of the proposed method. Runtimes for generating training sets and training the RF regression models for gates of the standard cell library is reported in Table 3. The reported runtimes in this table are associated with generating training sets with 2000 data. In order to improve the RF regression model (in more practical applications) it is justified to increase the size of the training set, resulting to more runtimes. While considerable runtime is required for training the ML-based model in the proposed method, there is no additional computation runtime in [11]. Once more, it should be emphasized that the training process is performed only one time and offline, therefore, it does not affect the timing analysis runtime of the proposed method.

**Table 7.** Runtime of developing RF Model

| Task | Runtime | Notes |
|---|---|---|
| Data Generation (HSPICE, Python) | 325 (hours) | • HSPICE simulations for all CL and TR simulation and all combinations of Vth per gate<br>• 3000 Monte Carlo iterations per gate |
| Model Training | 452 (minutes) | • Training<br>• BO optimization Hyperparameter tuning |
| Total | 332 (hours) | • All tasks performed offline<br>• no runtime impact |

## 5. Conclusion

As CMOS technology scales into the deep nanometer regime, the stochastic interplay between BTI and PV has evolved from a secondary effect into a primary reliability bottleneck. This work presents SMART, a comprehensive framework that resolves the longstanding conflict between computational scalability and predictive fidelity in reliability analysis. By effectively hybridizing ML with Monte Carlo simulations, we have established a methodology that bypasses the prohibitive runtime costs of traditional atomic model parameter extraction without compromising the statistical rigor required for safety-critical designs.

The experimental validation on ISCAS85 benchmark circuits demonstrates the transformative potential of this approach. SMART achieves an unprecedented 94.54% reduction in analysis time compared to state-of-the-art concurrent analysis methods, while maintaining an exceptionally low average accuracy error of 1.63%. This efficiency is driven by a strategic architectural decision to shift the computational complexity of atomic modeling to an offline training phase, enabling near-instantaneous runtime predictions of delay distributions (μ and δ). A critical innovation underlying these results is the integration of Bayesian Optimization (BO) for automated hyperparameter tuning. We demonstrated that replacing exhaustive grid searches with BO-driven probabilistic surrogates not only streamlines the model development process but significantly enhances the estimator's robustness across high-dimensional parameter spaces. This ensures that the RF models remain generalizable across diverse standard cell libraries and operating conditions.

In conclusion, SMART provides a scalable, high-speed alternative to commercial reliability tools that struggle with the volume of stochastic data in modern VLSI systems. Future work will extend this probabilistic framework to non-planar technologies, specifically FinFET and Gate-All-Around (GAA) architectures, and investigate the integration of real-time workload adaptation to further extend circuit operational lifetimes.